\documentclass[sigconf]{acmart}

\renewcommand\footnotetextcopyrightpermission[1]{}
\usepackage{booktabs}
\usepackage{multirow}
\usepackage{array}
\usepackage{graphicx}
\usepackage{balance}
\microtypesetup{expansion=false}
\AtBeginDocument{%
  }

\acmConference[ICAIF '26 Workshop]{LLMs and Generative AI for Finance (ICAIF 2026)}{November 14, 2026}{Milan, Italy}
\acmYear{2026}
\copyrightyear{2026}
\acmDOI{}
\acmISBN{}

\title{Honest Small Decision Models for Finance: Scale, Specialization Collapse, and Mixed-Corpus Remediation}

\author{Ajitabh Kumar}
\affiliation{%
  \institution{IntelCast}
  \city{Mumbai}
  \country{India}}

\begin{abstract}
Financial workflows need fast, calibrated, auditable decisions, not just generated text. We train \emph{System-1 decision models}, small language models fine-tuned to answer typed questions (\textsc{choice}/\allowbreak \textsc{score}/\allowbreak \textsc{noul}) with a probability in a single forward pass, across two model families (K2-Horizon and Qwen3.5/3.8) and nine rungs from 0.8B to 32B parameters, using one frozen LoRA recipe on a public decision benchmark and two new finance corpora: a programmatic-gold policy corpus (rule-reading with exact labels) and a reality-labeled outcome corpus (19{,}640 realized forward-return questions in walk-forward splits, plus a sealed 2026 test of 1{,}360 questions). Narrow fine-tuning costs general decision competence 9--45~percentage points, a data-mixture effect that shrinks with model scale and is associated with narrow-gradient volume. On reality-labeled questions, every trained rung from 0.8B to 32B lands at the horizon-conditioned prior baseline (accuracy 0.37--0.38, no detectable conditional edge under our controls); the models respond with flat, well-calibrated probabilities that track per-horizon base rates: naturally honest uncertainty, with a mid-scale calibration sweet spot (ECE 0.026 at 2B, degrading to 0.043--0.044 at 9--32B under prior drift). A mixed-corpus remediation carries all competencies in one adapter at modest cost: it matches or slightly exceeds the generalist specialist on the public benchmark while holding the policy ceiling and near-specialist outcome calibration, yielding a deployable artifact with per-domain temperature regimes and a band-dispatch escalation analysis: sharp, dispatchable bands where rules determine answers; flat, review-routing bands where reality does not.
\end{abstract}

\keywords{decision models, calibration, LoRA fine-tuning, model risk management, financial NLP, system-one models}

\begin{document}
\maketitle

\section{Introduction}
\label{sec:intro}

A trading desk, a compliance team, and a credit officer all need the same primitive: given a state and a typed question, a \emph{probability}, not a paragraph. Recent ``System-1'' decision models~\cite{typesafe} make this primitive architectural: the model consumes a state and typed questions (\textsc{choice} over options, \textsc{score} on an ordinal scale, \textsc{noul} returning $P(\mathrm{yes})$) and emits calibrated probabilities in one forward pass, at latencies and sizes suitable for production gates. Production deployments exist at exactly this shape (Cloudflare's \textsc{clef} is a 27B decision model post-trained from Qwen3.8-27B~\cite{clef}; the this-that model serves typed decisions in 30\,ms at a millionth of a cent~\cite{thisthat}), and a growing open ecosystem (\textsc{imajev}~\cite{imajev}, \textsc{strands-decider}~\cite{strands}, \textsc{decider}~\cite{decider}, \textsc{kev}~\cite{kev}) publishes recipes and leaderboards.

But three questions matter before an institution fine-tunes its own small decision layer, and the open literature answers none of them for the finance domain. \textbf{(1)~What happens to probability quality when the labels come from reality?} Decision-model training sets use clean or programmatically-solved labels; financial outcomes are noisy, regime-drifting, and close to unpredictable from standard technical features. \textbf{(2)~What does domain fine-tuning destroy?} Narrow SFT is known to erode general capabilities in decision models~\cite{imajev}; the governance question is how much, at what scale, and whether it can be fixed. \textbf{(3)~How should calibrated probabilities be turned into actions?} The gap between $P(\mathrm{yes})$ and an approve/review/reject dispatch is business logic, and needs a measurement discipline~\cite{pijev}.

This paper answers all three with a controlled study: one frozen training recipe, two model families, nine scale rungs (0.8B--32B), one public decision benchmark, and two new finance corpora built for this work. Our contributions:

\begin{itemize}
\item \textbf{A specialization-gap trend}: policy-only SFT costs 9--23pp of general decision accuracy, outcome-only SFT 15--45pp; damage shrinks roughly monotonically with scale (rung-to-rung wobble inside our $\sim$2pp resolution; at the top rungs 8.7--20.8pp remains) and tracks narrow-gradient volume (outcome training, with $8\times$ the rows, costs $1.7$--$3.8\times$ more at matched rungs, mean ${\approx}2.3\times$) A three-seed replication at Qwen-2B reproduces every headline.
\item \textbf{A remediation}: a single mixed-corpus adapter carries all competencies at modest cost, matching or slightly exceeding the generalist specialist on the public benchmark while holding the policy ceiling and near-specialist outcome calibration, with distinct per-domain temperatures in one artifact.
\item \textbf{An empirical characterization of the outcome task}: across all nine rungs and both families, trained accuracy sits at the horizon-conditioned prior baseline (the ``marginal ceiling''), classical controls detect no conditional signal on this feature set, and the models respond with flat, well-calibrated probabilities: honest uncertainty, not failure.
\item \textbf{A deployment analysis} turning probabilities into escalation policy via band dispatch: where rules determine answers the bands are empty and sharp (auto-decision is defensible within this corpus's regime); where reality does not, probabilities stay overwhelmingly in the middle band, and the honest default routes to review.
\item \textbf{A frozen, reproducible recipe} for LoRA decision fine-tuning, validated by a pre-registered hyperparameter probe on two families (field-consensus $10^{-4}$/r16 beats inherited agentic-program settings by $+2.6$pp accuracy on both families, $p{\le}0.006$), and an evaluation discipline with cluster-robust paired tests, foreign-panel regression checks, and leakage lints.
\item \textbf{Two finance corpora}: a programmatic-gold \emph{policy} corpus (2{,}651 rows, contrastive pairs, structure-keyed splits) teaching rule-reading; and a reality-labeled \emph{outcome} corpus (19{,}640 rows from live market data across walk-forward question-date splits, plus a sealed 1{,}360-question 2026 test; horizons 1--21 sessions; boundary-crossing labels audited) whose labels are realized forward returns.
\end{itemize}

The training recipe, corpus-construction procedures, and evaluation protocol are fully specified in this paper for independent replication; artifacts are available from the author on request. The sealed 2026 test split remains unevaluated.

\section{Background and Related Work}
\label{sec:related}

\paragraph{System-1 decision models.}
TypeSafe's Jev established the primitive set we adopt verbatim~\cite{typesafe}: \textsc{noul} returns $P(\mathrm{yes})$ for propositions (never a graded ``medium''), \textsc{score} grades on an ordered scale, \textsc{choice} distributes over categories. We read these probabilities from the language-model head over single-token codes under a pinned prompt contract (\S\ref{sec:setup}), the same readout used by the strongest open models: \textsc{imajev} (Qwen3.5 2B--9B, 255-code readout, LoRA $r{=}16$, $\alpha{=}32$) tops the JevBench board~\cite{imajev}; \textsc{strands-decider} replaces the head with a pointer head and documents pre-registered retrains with $\sigma{\approx}1.4$pp~\cite{strands}; \textsc{clef} is the production 27B instance on the same backbone family we test at the top rung~\cite{clef}. \textsc{decider}~\cite{decider} and \textsc{kev}~\cite{kev} contribute the programmatic-gold corpus pattern our policy corpus follows, and the temperature-scaling discipline (fit on held-out data by likelihood) our evaluation inherits; the field consensus is $T{>}1$ on clean-label decision data, and our reality-labeled rungs pull sharply toward 1 in an informative direction (\S\ref{sec:outcome}). A training-free line corrects the readout rather than the weights: AnyJev divides out a label prior estimated from unlabeled inputs and averages over option rotations with no gradient steps~\cite{zhang2026anyjevtechnicalreport}, and capable backbones read out competitively before any training~\cite{li2026llmasjevllmsjevstyledecision}. Our zero-shot grid bounds what that buys in finance: the best base reads policy rules at 0.812 where trained adapters sit at 0.99 or above, and no base model at any size beats chance on forward-direction questions. Domain evaluations are emerging in other regulated fields, e.g.\ clinical factuality~\cite{jevradiology}; we provide the finance one. Comparisons against full LLM judges find the decision readout cheaper and faster, and wrong in the same places~\cite{jevvllm}: shared failure modes, honestly priced.

\paragraph{LLMs and market prediction.}
Frontier LLMs extract some directional signal from \emph{text} such as news headlines~\cite{lopezlira}, and agentic finance systems push the generative stack toward autonomous deep research~\cite{xiao2026financeharness,wang2026mintagent}, but tabular technical states are a different regime: we show a logistic-regression control on our features recovers no signal, and no base model at any size (0.8B--32B) beats chance on our outcome questions. Our claim is deliberately not alpha; it is \emph{calibration}: small decision models trained on realized returns track the marginals closely and report them honestly.

\paragraph{Capability erosion.}
\textsc{imajev} reports that easy-decision SFT collapsed hard-item accuracy (64.9$\to$42.3) until a hard-question stage restored it~\cite{imajev}; the same erosion resembles the catastrophic-forgetting failure mode of neural networks~\cite{kirkpatrick2017}, documented for LLM continual fine-tuning~\cite{luo2025forgetting}; however, our finance adapters are fresh LoRA branches from the same base rather than continued training, so we measure a data-\emph{mixture} gap, not sequential forgetting. Independent work reports the same phenomenon on general-domain tasks: fine-tuning yields targeted gains with diminishing-to-negative transfer on capable backbones, and its losses are anchored by KL penalties to the base model~\cite{li2026llmasjevllmsjevstyledecision}, a loss-space remedy complementary to our data-space concatenation below. We characterize it as a two-axis trend (scale, narrow-gradient volume) across two families and two finance corpora, and show plain corpus concatenation is a sufficient remediation.

\section{Experimental Setup}
\label{sec:setup}

\paragraph{Models and families.}
Nine rungs across two families that enter the study as equals. \textbf{Qwen3.5} (Alibaba) at 0.8B/2B/4B/9B~\cite{qwen}, with Qwen3.8-27B~\cite{qwen38} as its top rung (the \textsc{clef} backbone; the Qwen arm switches generation 3.5$\to$3.8 at this rung only, documented as a confound, so within-Qwen scale reads stop at 9B). \textbf{K2-Horizon} (IFM) at 0.9B/3.7B/7B/32B~\cite{k2}: four dense members of a six-model fleet released 2026-09 as a fully open family (Apache 2.0 weights, checkpoints, data recipes, training code, and logs), with ${\sim}22$T pretraining tokens, reasoning trajectories in the mixture, agentic post-training, and vendor-reported state of the art at each size class. The two top rungs (Qwen3.8-27B vs K2-32B) form a class-4 pair that is size-mismatched by 5B with the generation switch only on the Qwen side; we compare them with cluster-robust tests and mark the confound where it matters.

\paragraph{Prompt contract and readout.}
Each question renders through the model's native chat template with thinking suppressed (K2 \texttt{reasoning\_effort=low}; Qwen \texttt{enable\_thinking=False}); the assistant region is an empty think block plus a single answer code. Codes: \textsc{choice} = letters A--Z, \textsc{score} = ordinal digits 0--9, \textsc{noul} = \texttt{true}/\texttt{false}, all verified single tokens per tokenizer, with per-model rendering pinned (a 24/24 render check including the 27B). The decision readout is a restricted softmax over the presented option tokens at the code position; the product is the full distribution, and argmax is used only inside accuracy metrics.

\paragraph{Frozen recipe.}
LoRA~\cite{hu2021lora} $r{=}16$, $\alpha{=}32$, dropout 0.05, on all language projections (Qwen includes its DeltaNet projections); lr $10^{-4}$ cosine, 3\% warmup, effective batch 8, two epochs with checkpoints at exact epoch boundaries; loss on the code token only. The recipe was frozen after a pre-registered probe: field-consensus $10^{-4}$/r16 vs.\ the inherited agentic-program $2{\cdot}10^{-4}$/r32 improved dev Brier 0.354$\to$0.330 and accuracy 0.754$\to$0.780 on K2-0.9B (McNemar~\cite{mcnemar1947}, $p{=}0.001$) and 0.375$\to$0.339 / 0.745$\to$0.771 on Qwen-0.8B ($p{=}0.006$); the same setting was independently converged on by \textsc{imajev} for its 2B.

\paragraph{Evaluation discipline.}
Multiclass Brier, NLL, ECE~\cite{guo2017} (10 equal-width bins on max-probability), AURC and coverage at error budgets; temperatures fitted by minimum NLL on a calibration split, verified with 5-fold spreads; cluster-robust paired tests (sign-flip over state-signature clusters); a foreign evaluation panel (an out-of-distribution decision suite) rides along with every artifact to catch silent erosion; leakage lints (cross-split signatures, label-space contract conformance, NaN/inf state fields) run at corpus freeze.

\section{Finance Decision Corpora}
\label{sec:data}

\paragraph{Public benchmark (``kev'').}
A 14{,}576-row public decision corpus (train) with 1{,}088 calibration and 1{,}352 development rows, plus a 764-row foreign development panel; used to validate the recipe and as the general-competence reference throughout.

\paragraph{Policy corpus (programmatic gold).}
2{,}651 records over three families (regime \textsc{choice}, risk \textsc{score}, eligibility \textsc{noul}): a policy with thresholds is drawn per structure; states draw feature values around the thresholds; labels are \emph{solved by code} under the stated rule; 251 contrastive pairs flip the deciding feature and assert the label flips; 1--3 distractor fields must be ignored. Splits are by \emph{structure} (14/\allowbreak 3/\allowbreak 3 per family = 1{,}857/\allowbreak 395/\allowbreak 399 train/\allowbreak calibration/\allowbreak development), so held-out splits carry unseen phrasings, thresholds and option orders; this is a rule-generalization read, never evidence of predictive skill. A base row (no adapter) measures what stock models do zero-shot.

\paragraph{Outcome corpus (reality-labeled).}
40 liquid US equities, daily bars 2016--present from a live market-data proxy, cached locally. Per sampled (symbol, session): a point-in-time state (RSI-14 and ADX-14, both in Wilder form~\cite{wilder}; close/SMA50, volume ratio, 5/21-day returns, 52-week distance) and one typed question with horizon $h\in\{1,5,10,21\}$: \textsc{noul} ``will close be higher after $h$ sessions?'', \textsc{score}/\textsc{choice} grading the realized close-to-close return into five bins with thresholds frozen from \emph{train-period} quantiles ($-2.23/-0.22/+1.15/+3.73\%$) and stated in the prompt. Labels are the realized returns. Splits are walk-forward by question date (states are point-in-time~\cite{xiao2026financeharness}): train $\le$2023 (15{,}640), calibration 2024 (2{,}000), development 2025 (2{,}000), test 2026+ (1{,}360, \textbf{sealed}: it ships unevaluated and is not opened in this paper). A question's forward window may cross its split's year boundary (70/69/59 rows in train/calibration/development, ${\le}3.5\%$); excluding the development crossers shifts accuracy and ECE by ${\le}0.002$ at every rung, so reported values stand.

\paragraph{Controls.}
(i)~\emph{Prior baseline}: the most frequent label per (qtype, horizon) from train, evaluated on development. (ii)~\emph{Logistic regression}: standardized, ridge-regularized, on the same seven noul features and rows. (iii)~Momentum sign. These convert ``our model found no edge'' into a statement about the data.

\section{Results on the Public Benchmark}
\label{sec:ladder}

Table~\ref{tab:ladder} condenses the nine-rung ladder. Each row is the rung's headline artifact (the ckpt column discloses the checkpoint; selection balances in-domain accuracy against foreign calibration: the ${\le}2$B rows favor epoch-2 dev, the 4B+ Qwen rows the epoch-1 boundary). Epoch 2 erodes foreign-panel ECE at 8 of 9 rungs (K2-0.9B excepted, where it weakly dominates on both panels); at Q-2B it improves foreign accuracy and Brier while still eroding ECE ($0.060{\to}0.086$); this is a robust pattern we return to in \S\ref{sec:deploy}. Three findings. \textbf{(1) No detectable family advantage:} prior agentic RL-distillation experiments in our stack found K2 dramatically more plastic than Qwen, but across four matched-scale comparisons at epoch-2 artifacts run in the same window, plus the class-4 pair under a common epoch-1 rule (cluster-robust tests), the decision regime shows no consistent ordering (e.g.\ at 3.7/4B, dev $p{=}0.73$, transfer $p{=}0.63$; at 27/32B under the common rule, dev $p{=}0.90$). The class-4 comparison is checkpoint-sensitive: a common epoch-2 rule shows a weak, non-significant Qwen trend ($p{=}0.07$), and the headline artifacts sit between ($p{=}0.57$); no rule yields a significant ordering. Family choice is deployment freedom, not a research finding here. \textbf{(2) Scale curves differ in timing:} K2 stalls between 3.7B and 7B (0.825/0.825) and climbs again at 32B (0.837), while Qwen climbs earlier, through 27B, to the program best (0.848 acc / 0.243 Brier); epoch-1-boundary artifacts hold the best transfer on both sides (27B: 0.791/0.281/0.030; K2's best is 32B@1822 at 0.754/0.314/0.037). \textbf{(3) Boundary artifacts dominate:} at 8 of 9 rungs the epoch-1 checkpoint carries the better-foreign-calibration artifact (K2-0.9B excepted); at Q-2B epoch 2 improves foreign accuracy and Brier but not calibration. Training longer usually buys in-domain accuracy at the cost of out-of-domain calibration.

\begin{table}[t]
\centering
\small
\caption{Public-benchmark ladder, two epochs, post-temperature, rows ascending in size. Rows are per-rung headline artifacts balancing in-domain and foreign quality (ckpt column discloses the choice; the K2 rows and ${\le}2$B rows favor epoch-2 dev, the 4B+ Qwen rows the epoch-1 boundary). Q3.8-27B switches Qwen generation 3.5$\to$3.8 at this rung only (\S\ref{sec:setup}). Dev = in-domain (n=1{,}352); Tr = foreign panel (n=764). Recipe frozen across all rungs.}
\label{tab:ladder}
\begin{tabular}{llcccc}
\toprule
Rung & ckpt & Dev acc & Dev Brier & Tr acc & Tr ECE \\
\midrule
Q-0.8B & 3644 & 0.801 & 0.317 & 0.581 & 0.111 \\
K2-0.9B & 3644 & 0.783 & 0.322 & 0.611 & 0.078 \\
Q-2B & 3644 & 0.808 & 0.295 & 0.649 & 0.086 \\
K2-3.7B & 3644 & 0.825 & 0.268 & 0.711 & 0.080 \\
Q-4B & 1822 & 0.825 & 0.260 & 0.707 & 0.069 \\
K2-7B & 3644 & 0.825 & 0.272 & 0.728 & 0.050 \\
Q-9B & 1822 & 0.826 & 0.258 & 0.740 & 0.050 \\
Q3.8-27B & 1822 & 0.831 & 0.254 & \textbf{0.791} & \textbf{0.030} \\
K2-32B & 3644 & 0.837 & 0.250 & 0.747 & 0.066 \\
\bottomrule
\end{tabular}
\end{table}

\section{Finance Results}
\label{sec:finance}

\subsection{Zero-shot grid: what stock models know}
\label{sec:zeroshot}

Running every \emph{untrained} base through the same readout isolates the fine-tuning contribution (Table~\ref{tab:zeroshot}). Rule-reading is a base capability that scales on the Qwen side (0.383$\to$0.812); K2 peaks near 0.58 at 3.7--7B and falls to 0.43 at 32B, so base competence is not monotone in size. Fine-tuning closes the entire gap to ceiling at every rung. Forward-direction questions are different: \emph{no} base model at any size beats chance (0.29--0.32 vs.\ 0.304), and base probabilities about forward moves are badly miscalibrated (ECE 0.14--0.36). The outcome experiments below are therefore a calibration study against prior baselines, not an accuracy race.

\begin{table}[t]
\centering
\small
\caption{Zero-shot (base, same readout) vs.\ trained adapters. Chance: policy 0.331, outcome 0.304.}
\label{tab:zeroshot}
\begin{tabular}{lcccc}
\toprule
 & \multicolumn{2}{c}{Policy} & \multicolumn{2}{c}{Outcome} \\
Rung & base & adapter & base & adapter \\
\midrule
Q-0.8B & 0.383 & 0.995 & 0.304 & 0.371 \\
K2-0.9B & 0.388 & 0.992 & 0.323 & 0.373 \\
Q-2B & 0.589 & 1.000 & 0.299 & 0.383 \\
K2-3.7B & 0.564 & 1.000 & 0.291 & 0.382 \\
Q-4B & 0.654 & 1.000 & 0.294 & 0.379 \\
K2-7B & 0.584 & 1.000 & 0.299 & 0.378 \\
Q-9B & 0.747 & 1.000 & 0.311 & 0.382 \\
Q3.8-27B & 0.812 & 1.000 & 0.302 & 0.378 \\
K2-32B & 0.429 & 1.000 & 0.308 & 0.370 \\
\bottomrule
\end{tabular}
\end{table}

\subsection{Policy cell: sharp where rules decide}
\label{sec:policy}

Trained adapters hit 0.992--1.000 on held-out structures, near-ceiling generalization across unseen phrasings, thresholds and option orders, with the loss curve collapsing within one epoch (programmatic gold is exactly learnable; the corpus is a validation instrument, not a discriminator). The interesting policy-cell result is operational, in \S\ref{sec:deploy}: the learned probabilities are sharp enough to drive dispatch bands with measured, empty gaps.

\subsection{Outcome cell: flat and honest where reality decides}
\label{sec:outcome}

Table~\ref{tab:outcome} is the paper's core finance table. Four findings.

\paragraph{(1) Accuracy remains close to the horizon-conditioned prior baseline.}
All nine rungs, both families, land at 0.370--0.383 (prior baseline 0.379); the directional (\textsc{noul}) subtask never exceeds the always-up baseline meaningfully (0.501--0.521 vs.\ 0.519). By question type, the adapters show no demonstrated advantage over the prior baseline: the per-(qtype, horizon) prior table from train already achieves noul 0.519 / score 0.303 / choice 0.305, matching the adapters. Throughout, ``marginal ceiling'' is shorthand for this operational fact: no tested model, family, or size demonstrates accuracy above the horizon-conditioned prior baseline under our controls.

\paragraph{(2) No conditional signal under this protocol.}
Logistic regression on the same features and rows: dev accuracy 0.514 vs.\ always-up 0.519, dev Brier 0.2510 \emph{worse than the constant base-rate predictor} (0.2496), all $|w|\le 0.08$ after standardization; momentum sign 0.504; a nonlinear gradient-boosted control on the same features is no better (accuracy 0.496 vs.\ 0.519 always-up, Brier 0.282). Under this protocol (seven technical features, a linear control, horizons of 1--21 sessions), no conditional directional signal is detectable for these large caps. The tested models show no established directional advantage over the prior-based controls, and base models of every size sit at chance.

\paragraph{(3) The honest response is flat, calibrated probability.}
Temperature fits land near 1.0--1.4 (vs.\ 1.6--2.0 on clean-label corpora across the field): training against noisy realized returns leaves little overconfidence for temperature to correct. Mean $P(\mathrm{up})$ tracks the per-horizon realized up-rate (e.g.\ 2B: 0.523/0.520, 0.535/0.534, 0.546/0.539 at $h{=}1/5/10$); the model has learned each horizon's base rate and little else, and says so. A per-(qtype,horizon) prior table from train is better calibrated on pooled ECE (0.019 vs.\ 0.026 for the 2B rung) while the 2B adapter has slightly higher accuracy and lower Brier and NLL (0.383 vs.\ 0.379, 0.666 vs.\ 0.673, 1.206 vs.\ 1.224; Table~\ref{tab:priorscores}), so pooled ECE alone cannot establish probability quality, and neither can proper scores alone: they mix discrimination with calibration, we do not decompose them here, and under date-block resampling (common-date blocks, temporal dependence) the specialist's proper-score advantage over the table is itself unresolved (Brier $+0.005$, 95\% CI $[-0.001,+0.010]$; NLL $+0.005$, $[-0.001,+0.011]$). Calibration is also heterogeneous by question type: the 2B adapter's score questions carry ECE 0.050 against 0.026 pooled.

\paragraph{(4) The calibration sweet spot is mid-scale under drift.}
Outcome ECE is U-shaped in scale: 0.040/\allowbreak 0.034/\allowbreak \textbf{0.026}/\allowbreak 0.030/\allowbreak 0.035/\allowbreak 0.036/\allowbreak 0.044/\allowbreak 0.043/\allowbreak 0.043 from 0.8B to 32B, with 2B the lowest measured pooled ECE in this evaluation and the top rungs flat at 0.043--0.044. A consistent reading: larger models track train-period marginals more sharply, so the 2025 prior drift (e.g.\ 21-session up-rate 0.555 in train vs.\ 0.482 realized in development) hits them harder; at 9B, mean $P(\mathrm{up})$ is 0.629 vs.\ 0.534 realized at $h{=}5$ and 0.670 vs.\ 0.482 at $h{=}21$ (why the sharper tracking arises is not isolated here). Temperature cannot repair a shifted center. Under walk-forward drift, the smaller rungs measured the best probability quality in this evaluation, a counter-intuitive deployment observation; a rolling-marginal recalibration belongs in the gate, not the model.

\begin{table}[!t]
\centering
\small
\caption{Proper scores vs.\ pooled ECE on the outcome development set (n=2{,}000; adapters temperature-scaled). The specialist has slightly higher accuracy and lower Brier and NLL despite higher pooled ECE; the mixed adapter returns that advantage to the prior table's level. Proper-score differences vs.\ the prior table are within date-block bootstrap noise (\S\ref{sec:outcome}).}
\label{tab:priorscores}
\begin{tabular}{lcccc}
\toprule
Predictor & Acc & Brier & NLL & ECE \\
\midrule
Train prior table & 0.379 & 0.673 & 1.224 & \textbf{0.019} \\
Q-2B outcome specialist & \textbf{0.383} & \textbf{0.666} & \textbf{1.206} & 0.026 \\
Q-2B mixed adapter & 0.376 & 0.673 & 1.223 & 0.044 \\
\bottomrule
\end{tabular}
\end{table}

\begin{table}[t]
\centering
\small
\caption{Outcome cell (development 2025, n=2{,}000; epoch-2 finals). ``Collapse'' = general-competence loss on the public benchmark after narrow training (\S\ref{sec:collapse}).}
\label{tab:outcome}
\begin{tabular}{lccccc}
\toprule
Rung & Acc & ECE & $T$ & \textsc{noul} & Collapse \\
\midrule
Q-0.8B & 0.371 & 0.040 & 1.06 & 0.511 & $-$45.3 \\
K2-0.9B & 0.373 & 0.034 & 1.06 & 0.521 & $-$44.3 \\
Q-2B & 0.383 & \textbf{0.026} & 1.14 & 0.519 & $-$40.0 \\
K2-3.7B & 0.382 & 0.030 & 1.06 & 0.519 & $-$37.6 \\
Q-4B & 0.379 & 0.035 & 1.10 & 0.519 & $-$40.3 \\
K2-7B & 0.378 & 0.036 & 1.18 & 0.519 & $-$23.2 \\
Q-9B & 0.382 & 0.044 & 1.38 & 0.518 & $-$19.9 \\
Q3.8-27B & 0.378 & 0.043 & 1.33 & 0.501 & $-$14.5 \\
K2-32B & 0.370 & 0.043 & 1.23 & 0.518 & $-$20.8 \\
\midrule
\multicolumn{6}{l}{Prior baseline 0.379; chance 0.304; always-up 0.519.} \\
\bottomrule
\end{tabular}
\end{table}

\begin{figure*}[t]
\centering
\includegraphics[width=0.85\textwidth]{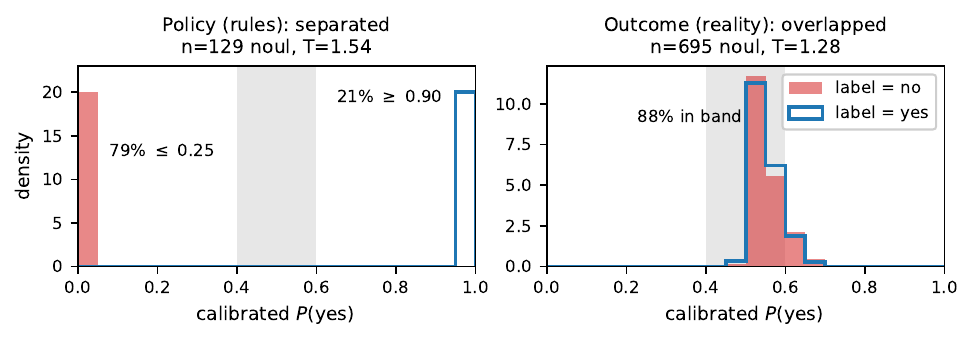}
\caption{Calibrated $P(\mathrm{yes})$ by realized label for the Q-2B mixed adapter, post domain temperature (policy $T{=}1.54$, outcome $T{=}1.28$), common axis. Left: policy development noul questions (n=129) separate into sharp extremes (21\% at $P{\ge}0.90$, 79\% at $P{\le}0.25$). Right: outcome development 2025 noul questions (n=695) overlap near the middle; the shaded $[0.40,0.60)$ region is the \emph{proposed} review band, covering 88\% of outcome mass.}
\label{fig:bands}
\end{figure*}

\section{Specialization Collapse and Remediation}
\label{sec:collapse}

\begin{figure*}[t]
\centering
\includegraphics[width=0.85\textwidth]{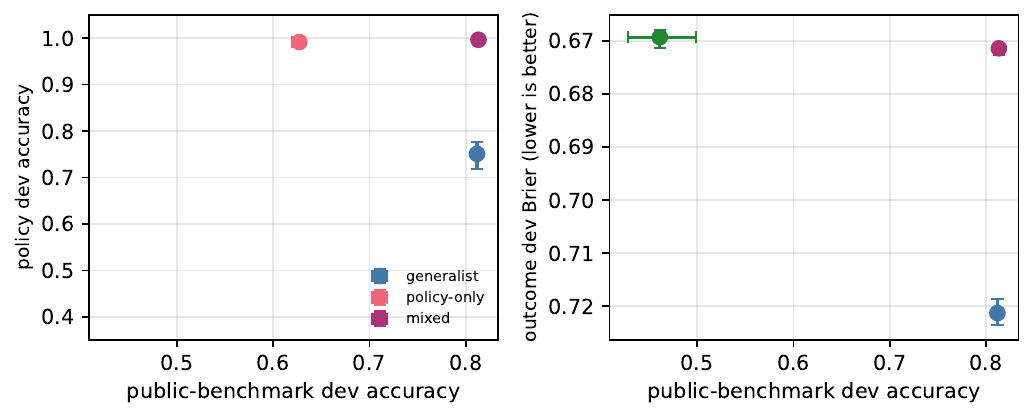}
\caption{The specialization trade-off and its remediation at Qwen-2B (three seeds; mean with min--max whiskers, Table~\ref{tab:replication}). Left: policy development accuracy against public-benchmark accuracy; right: outcome development Brier (lower is better) against public-benchmark accuracy. Narrow training buys one finance axis at a large cost in general competence (and leaves the untrained finance axis at chance); mixed training recovers the generalist position on the public benchmark while holding both finance axes near their specialists.}
\label{fig:tradeoff}
\end{figure*}

Fine-tuning on one finance corpus damages general decision competence, measured by re-evaluating each narrow adapter on the public benchmark (Table~\ref{tab:outcome}, last column; policy-training analog in Table~\ref{tab:mixed} caption). Throughout, this is a gap between \emph{separately trained} adapters from the same base, a data-mixture effect rather than sequential forgetting; the mixed run below shows the capacity is intact. Three regularities. \textbf{Scale buffers:} policy-training damage roughly halves from $-23.3$/$-20.8$pp at 0.8/0.9B to $-8.7$pp at 27B (monotone to plateau, with rung-to-rung wobble inside our $\sim$2pp resolution); outcome-training damage falls from $-45$pp to $-14.5$pp. \textbf{The larger corpus is associated with the larger gap:} at matched rungs, outcome training (15{,}640 rows vs.\ 1{,}857) costs $1.7$--$3.8\times$ more general competence (mean ${\approx}2.3\times$); this is an association, not an isolated mechanism, because label noise, task difficulty and step count change together with corpus size. \textbf{Both families behave alike.} A three-seed replication at 2B (Table~\ref{tab:replication}, Figure~\ref{fig:tradeoff}) confirms the ordering and sizes the run-to-run spread: policy-only collapse $-18$ to $-19$pp, outcome-only $-31$ to $-38$pp, the mixed adapter within ${\sim}1$pp of the generalist on every panel. The converse of the gap also holds: the kev-only generalist sits at chance on outcome questions (accuracy 0.300, ECE 0.092 across seeds); finance competence arrives only with finance data. The governance reading: a compliance fine-tune silently forfeits general decision competence unless the training mixture is designed; the damage tracks (scale, narrow-gradient volume) as an empirical association (independently reported for general-domain tasks~\cite{li2026llmasjevllmsjevstyledecision}).

\paragraph{Remediation: one adapter, all competencies.}
Concatenating all three corpora (32{,}073 rows) under the same recipe and evaluating the single adapter on every panel meets and exceeds pre-registered criteria (Table~\ref{tab:mixed}; three-seed replication in Table~\ref{tab:replication}): both mixed adapters match or slightly exceed their generalist-only specialists on the public benchmark ($+0.6$/$+0.9$pp, point estimates within our $\sim$2pp resolution) while holding the policy ceiling and the outcome marginal ceiling, at modest, measurable cost (foreign-panel accuracy $-0.9$/$-1.1$pp; outcome ECE $0.026{\to}0.044$ at 2B and $0.030{\to}0.041$ at 3.7B; and at 2B the specialist's Brier/NLL advantage over the prior table returns to table level, Table~\ref{tab:priorscores}). Each carries distinct per-domain temperatures in one artifact: kev $\approx$1.8, policy pinned at ceiling, outcome $\approx$1.1--1.3. The collapse is a data-composition artifact, not a capacity limit; plain concatenation substantially closes it. This is the deployable object: sharp policy bands, honest outcome probabilities, general competence, one checkpoint.

\begin{table}[t]
\centering
\small
\setlength{\tabcolsep}{3.5pt}
\caption{Mixed-corpus remediation (concat train, frozen recipe), demonstrated at the two tested sizes. Specialist references in parentheses. Policy-only training collapse for these rungs: $-17.5$ (Q-2B), $-13.9$ (K2-3.7B).}
\label{tab:mixed}
\begin{tabular}{lcccccc}
\toprule
Adapter & Dev & Tr & Policy & Out acc & Out ECE & $T$ k/p/o \\
\midrule
Q-2B mix & \textbf{0.814} & 0.640 & 0.995 & 0.376 & 0.044 & \begin{tabular}[c]{@{}c@{}}1.79/1.54\\1.28\end{tabular} \\
 & (0.808) & (0.649) & & & & \\
K2-3.7B mix & \textbf{0.834} & 0.700 & 1.000 & 0.369 & 0.041 & \begin{tabular}[c]{@{}c@{}}1.79/0.02\\1.14\end{tabular} \\
 & (0.825) & (0.711) & & & & \\
\bottomrule
\end{tabular}
\end{table}

\section{From Probabilities to Actions: Band Dispatch}
\label{sec:deploy}

A decision model returns probabilities; action thresholds are business logic layered on top, derived from decision costs ($0.5$ exactly when the costs are symmetric) and validated on held-out data. Following the measurement doctrine of \textsc{pi-jev}~\cite{pijev} (thresholds are \emph{measured}, not chosen: each sits in the empty band between the score distributions of positive and negative states, and must survive run-to-run drift), we evaluate dispatch bands on labeled data. Figure~\ref{fig:bands} shows the two regimes side by side for one artifact (the Q-2B mixed adapter) on a common probability axis. The model-side dual of this routing is AnyJev's capped escalation, which sends low-margin decisions to the model's own reasoning under a hard share cap~\cite{anyjev2026}; our bands route the flat regime to human review instead.

\paragraph{Sharp regime (policy questions).}
On held-out structures (Figure~\ref{fig:bands}, left), noul mass is bimodal at the extremes: 21\% of rows at $P\ge 0.90$ (n=27) with realized yes-rate \textbf{1.000}, 79\% at $P\le 0.25$ (n=102) with realized yes-rate \textbf{0.000}. The bands are empty of errors, but the counts are small: zero errors in 27 positive-band examples still admits a one-sided 95\% error bound of 10.5\% even under independence. Within this corpus's regime the bands are candidate operating policies: they have volume, calibrate within band, and the gate could dispatch most of this corpus; institutional adoption requires validating the bound on more data.

\paragraph{Pre-trade gate (controlled workflow simulation).}
We assemble the bands into a controlled simulation of a pre-trade policy gate: each policy-development \textsc{noul} question (n=129) is a proposed action asking eligibility; the adapter's calibrated $P(\mathrm{true})$ dispatches approve ($P{\ge}0.90$), reject ($P{\le}0.25$), or review; the programmatic gold label supplies the reference decision. All three adapters separate this gate with zero incorrect approvals and zero incorrect rejections at these bands (review rate 0--0.8\%; one-sided 95\% error bounds 10.5\%/2.9\% at these counts), including the kev-only generalist: rule-reading transfers from the public benchmark, and the gate itself is not the binding constraint. The workflow differential appears on the other panels: the policy-only specialist pays $-17.5$pp of general competence for a gate the generalist already passes (\S\ref{sec:collapse}), the generalist carries no outcome competence, and the mixed adapter holds the gate, the public benchmark, and near-specialist outcome calibration in one checkpoint. The separation is seed-robust: six independently seeded replications of the policy and mixed adapters produce the identical 27/102/0 split with zero errors. This is a simulation on programmatic-gold structures, not a production workflow evaluation.

\paragraph{Flat regime (outcome questions).}
On 2025 development data (Figure~\ref{fig:bands}, right), the Q-2B outcome specialist keeps 99.9\% of noul mass in $[0.40,0.60)$ (within-band gap $-0.026$); the mixed adapters sit at 83--88\%, and the 9B specialist at 57\%; the flat-regime reading holds at every rung measured, weakening with scale and mixing but never inverting. No empty band appears at any threshold worth having, and the probabilities agree by staying concentrated toward the middle (57--99.9\% in band across the measured adapters). Routing to review is the policy these statistics support: a chosen default under asymmetric review costs, not a property of the model. Two independent lines, deployment doctrine and calibrated probabilities, converge on the same operating posture. Where an institution needs automation on market-direction questions, the lever is richer states (cross-sectional ranks, fundamentals, news), not bigger models on the same features: our zero-shot and trained grids show no size extracts conditional signal from standard technical states.

\paragraph{Per-domain calibration in one artifact.}
The mixed adapter ships three temperature regimes; applying each is a serving-time concern, and the drift analysis (\S\ref{sec:outcome}) adds the operational rule: under walk-forward label drift, refit marginals on a rolling window, a gate concern, not retraining (the window matters: a one-year re-estimate of the prior table is noise-dominated and worse than the 2016--2023 table on every 2025 development metric, so use the longest stable window).

\begin{table}[t]
\centering
\footnotesize
\setlength{\tabcolsep}{3.5pt}
\caption{Three-seed replication at Qwen-2B (seeded LoRA initialization, frozen recipe otherwise): mean (min--max) over three independently seeded runs per condition. Foreign panel for reference: outcome-only 0.461 [.441--.479], policy-only 0.619 [.617--.620], mixed 0.625 [.611--.643]. Single-run artifacts of Tables~\ref{tab:ladder}--\ref{tab:mixed} sit inside or within ${\sim}1$pp of these ranges (the outcome-only collapse is the widest cell: $-31$ to $-38$pp vs.\ its seed baselines, single-run value $-40.0$ just beyond); the mixed adapter is the most seed-stable, within ${\sim}1$pp of the generalist on the public benchmark while carrying all finance competencies.}
\label{tab:replication}
\begin{tabular}{lccccc}
\toprule
Condition & Kev & Policy & Out acc & Out Brier & Out ECE \\
\midrule
Generalist & \begin{tabular}[c]{@{}c@{}}0.812\\{\tiny(.810--.813)}\end{tabular} & \begin{tabular}[c]{@{}c@{}}0.751\\{\tiny(.719--.777)}\end{tabular} & \begin{tabular}[c]{@{}c@{}}0.300\\{\tiny(.294--.308)}\end{tabular} & \begin{tabular}[c]{@{}c@{}}0.721\\{\tiny(.719--.724)}\end{tabular} & \begin{tabular}[c]{@{}c@{}}0.092\\{\tiny(.076--.101)}\end{tabular} \\
Policy-only & \begin{tabular}[c]{@{}c@{}}0.627\\{\tiny(.619--.632)}\end{tabular} & \begin{tabular}[c]{@{}c@{}}0.992\\{\tiny(.983--.998)}\end{tabular} & -- & -- & -- \\
Outcome-only & \begin{tabular}[c]{@{}c@{}}0.461\\{\tiny(.428--.498)}\end{tabular} & -- & \begin{tabular}[c]{@{}c@{}}0.378\\{\tiny(.378--.379)}\end{tabular} & \begin{tabular}[c]{@{}c@{}}0.669\\{\tiny(.668--.671)}\end{tabular} & \begin{tabular}[c]{@{}c@{}}0.034\\{\tiny(.028--.038)}\end{tabular} \\
Mixed & \begin{tabular}[c]{@{}c@{}}0.813\\{\tiny(.812--.814)}\end{tabular} & \begin{tabular}[c]{@{}c@{}}0.997\\{\tiny(.990--1.00)}\end{tabular} & \begin{tabular}[c]{@{}c@{}}0.374\\{\tiny(.367--.381)}\end{tabular} & \begin{tabular}[c]{@{}c@{}}0.671\\{\tiny(.670--.673)}\end{tabular} & \begin{tabular}[c]{@{}c@{}}0.044\\{\tiny(.041--.047)}\end{tabular} \\
\bottomrule
\end{tabular}
\end{table}

\section{Limitations}
\label{sec:limits}

(1) Ladder cells are single training runs; the family-ordering and collapse reads rest on item-level cluster-robust tests, not retrain replications at every rung. A three-seed replication at Qwen-2B with seeded LoRA initialization (Table~\ref{tab:replication}) reproduces every headline in sign and magnitude: the mixed remedy is seed-stable to ${\sim}1$pp on all panels, while the outcome-only collapse is the widest cell ($-31$ to $-38$pp across seeds; the single-run value $-40.0$ sits just beyond); differences under $\sim$2pp remain unresolved elsewhere. (2) With 695 noul rows the approximate 95\% binomial half-width is $\pm$3.7pp; small differences remain unresolved, and given temporal dependence this independence-based interval is only a rough precision reference. (3) The class-4 pair (Qwen3.8-27B vs K2-32B) is size-mismatched by 5B with the generation switch only on the Qwen side; under a common epoch-1 rule the cluster test shows no ordering ($p{=}0.90$), under a common epoch-2 rule a weak non-significant Qwen trend ($p{=}0.07$), between headline artifacts $p{=}0.57$: the checkpoint rule matters, and this is not a clean size-matched family comparison. (4) The outcome corpus is one universe (40 US large caps), one feature vocabulary, one period; prior drift is characterized, not solved; question windows overlap and share market exposure, so binomial intervals understate uncertainty; the retrospective boundary-crossing audit covers reported development metrics only, and its 70 train-period crossers extend labels into calibration's window; retraining on purged splits could shift fitted models and temperatures, and is not re-run here. (5) Our \textsc{noul} is forced-choice, with no abstention channel; an \texttt{unknown} code would sharpen the flat regime's routing. (6) The sealed 2026 test split remains unevaluated by design.

\balance
\section{Conclusion}
\label{sec:conclusion}

We trained the same frozen decision recipe across two families and nine rungs and put it to finance work. The recipe holds (no detectable family ordering, scale curves, boundary artifacts); reality-labeled finance questions expose what clean benchmarks cannot: accuracy at the prior baseline at every rung measured (the ``marginal ceiling''), near-1 temperatures, per-horizon base-rate tracking, and a mid-scale calibration sweet spot under drift; narrow specialization forfeits general competence, with the gap associated with scale and narrow-gradient volume; and concatenating corpora closes it in one deployable artifact. The resulting artifact is small, honest, and governable: sharp and dispatchable where rules decide, flat and escalating where reality does.

\appendix
\section{Artifact Specification}
\label{app:artifact}

\paragraph{Base models.}
Qwen3.5 at 0.8B/2B/4B/9B and Qwen3.8-27B (Hugging Face, Apache 2.0); K2-Horizon at 0.9B, 3.7B, 7B, and 32B (IFM, Apache 2.0; 32B snapshot \texttt{b89c0f9e}). The 27B and 32B rungs train with gradient checkpointing on (numerically neutral); all other settings are identical across rungs.

\paragraph{Training.}
LoRA $r{=}16$, $\alpha{=}32$, dropout 0.05, all language projections (Qwen including DeltaNet projections); lr $10^{-4}$ cosine, 3\% warmup, effective batch 8 (bs1 $\times$ ga8), two epochs with exact-boundary checkpoints; loss on the code token only; bf16; trainer seed 42 (framework default; LoRA initialization is not separately re-seeded); corpus generators are seeded (policy 42, outcome 11). Decision records are identical across rungs; rendering is pinned per model and hash-verified within each family (K2-3.7B/7B/32B share one render; K2-0.9B has its own tokenizer profile; shared K2 vocabulary 250{,}624 tokens).

\paragraph{Corpora.}
kev public suites from \texttt{jaredpalmer/kev-suites} (decision-v7 train/calibration/development; transfer-v4 foreign panel, eval-only), manifest-frozen (SHA-256, 2026-10-02); 1{,}000 train records with ${>}26$ options dropped at the contract cap. Policy corpus generated with seed 42 (three structure families, 251 contrastive pairs, structure-keyed splits). Outcome corpus: 40 liquid US large-cap equities, daily bars from 2016-01-01 to the cache date via an IntelCast market-data proxy (historical-price-eod, cached per symbol), split-adjusted closes (\texttt{adjClose} when present), point-in-time features (RSI-14/ADX-14 Wilder, close/SMA50, volume ratio, 5/21-day returns, 52-week distance), five-bin thresholds frozen from train-period quantiles ($-2.23/-0.22/+1.15/+3.73\%$), generator seed 11, walk-forward splits by question date (\S\ref{sec:data}).

\paragraph{Evaluation.}
Scorer v1.1: multiclass Brier, NLL, ECE (10 equal-width bins), AURC, coverage at error budgets; temperatures fitted by NLL minimization on calibration splits ($p^{(1/T)}$ kernel, 161 log-spaced candidates over 0.02--8.0); family tests are two-sided sign-flip permutation over state-signature clusters (10k permutations, seed 0); zero-shot grids use the same readout and panels as trained adapters.

\paragraph{Environment.}
Single RTX PRO 6000 (96GB); PyTorch with Transformers Trainer + PEFT; restricted-softmax readout in the evaluation harness.

\bibliographystyle{ACM-Reference-Format}
\bibliography{references}

\end{document}